\documentclass[letterpaper, 10 pt, conference]{ieeeconf}  % Comment this line out if you need a4paper

\IEEEoverridecommandlockouts                              % This command is only needed if 
\usepackage{amsmath} % assumes amsmath package installed
\usepackage{amssymb}  % assumes amsmath package installed
\usepackage{graphicx}   % 插入图片用
\usepackage{caption}    % 控制图题样式
\usepackage{booktabs, multirow}
\usepackage{setspace}

\title{\LARGE \bf
FishBEV: Distortion-Resilient Bird’s Eye View Segmentation with Surround-View Fisheye Cameras
}

\author{Hang Li$^{1,2}$, Dianmo Sheng$^{3}$, Qiankun Dong$^{1,2}$, Zichun Wang$^{1,2}$, Zhiwei Xu$^{4}$, Tao Li$^{1,2\ast}$% <-this % stops a space
\thanks{*This work was supported by the National Natural Science Foundation of China (62272248).}% <-this % stops a space
\thanks{$^{1}$College of Computer Science, Nankai University, Tianjin, 300350, China. 
        {\tt\small lihangnk@mail.nankai.edu.cn}, {\tt\small \{qiankund, wzc, litao\}@nankai.edu.cn}}%
\thanks{$^{2}$Key Laboratory of Data and Intelligent System Security, Ministry of Education, China}%
\thanks{$^{3}$School of Cyber Security, University of Science and Technology of China, Hefei, Anhui, 230026, P.R.China.
        {\tt\small dmsheng@mail.ustc.edu.cn}}%
\thanks{$^{4}$Haihe Lab of ITAI, Institute of Computing Technology, Chinese Academy of Sciences, Beijing, China.
        {\tt\small xuzhiwei2001@ict.ac.cn}}%
\thanks{$^\ast$Corresponding author: Tao Li (litao@nankai.edu.cn).}%
}

\begin{document}

\maketitle
\thispagestyle{empty}
\pagestyle{empty}

%%%%%%%%%%%%%%%%%%%%%%%%%%%%%%%%%%%%%%%%%%%%%%%%%%%%%%%%%%%%%%%%%%%%%%%%%%%%%%%%
\begin{abstract}

As a cornerstone technique for autonomous driving, Bird's Eye View (BEV) segmentation has recently achieved remarkable progress with pinhole cameras. However, it is  non-trivial to extend the existing methods to fisheye cameras with severe geometric distortion, ambiguous multi-view correspondences and unstable temporal dynamics, all of which significantly degrade BEV performance. To address these challenges, we propose FishBEV, a novel BEV segmentation framework specifically tailored for fisheye cameras. This framework introduces three complementary innovations, including a Distortion-Resilient Multi-scale Extraction (DRME) backbone that learns robust features under distortion while preserving scale consistency,  an Uncertainty-aware Spatial Cross-Attention (U-SCA) mechanism that leverages uncertainty estimation for reliable cross-view alignment, a Distance-aware Temporal Self-Attention (D-TSA) module that adaptively balances near field details and far field context to ensure temporal coherence. Extensive experiments on the Synwoodscapes dataset demonstrate that FishBEV consistently outperforms SOTA baselines, regarding the performance evaluation of FishBEV on the surround-view fisheye BEV segmentation tasks.

\end{abstract}

%%%%%%%%%%%%%%%%%%%%%%%%%%%%%%%%%%%%%%%%%%%%%%%%%%%%%%%%%%%%%%%%%%%%%%%%%%%%%%%%
\section{INTRODUCTION}

Bird's Eye View (BEV) segmentation plays a key role in autonomous driving and mobile robotics, providing a unified spatial representation for downstream tasks such as planning, navigation, and scene understanding \cite{p1, p2, p3}. In recent years, surround-view fisheye cameras have become an indispensable perception device in many autonomous driving systems due to their ultra-wide field of view, compact mounting structure, and low cost \cite{p4, p5}. However, the severe distortion and nonlinear projection in fisheye perspective view (PV) images pose significant challenges for accurate BEV segmentation, especially compared to standard pinhole or panoramic image systems \cite{p6, p7}. Therefore, designing effective algorithms to bridge the gap between the distorted observations of surround-view fisheye images and the spatially consistent BEV representation remains a major challenge.
\begin{figure}
    \centering
    \includegraphics[width=1\linewidth]{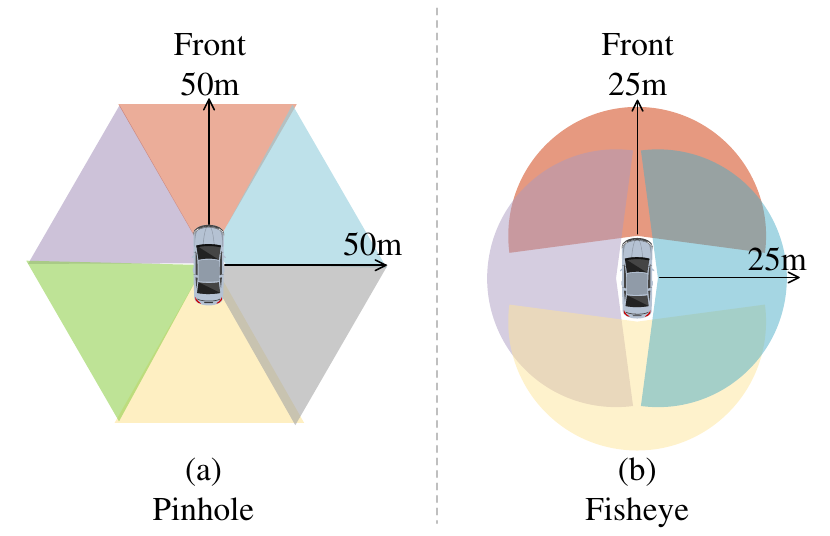}
    \caption{Comparison of BEV field-of-view coverage between pinhole cameras (left) and fisheye cameras (right). The pinhole configuration employs six cameras with non-overlapping $\sim$90$^\circ$ views, extending up to 50\,m, while the fisheye setup uses four wide-angle cameras with $\sim$180$^\circ$ views and significant overlap, covering up to 25\,m.}
    \label{pin_fish}
    \vspace{-0.8em}
\end{figure}

Current research on BEV segmentation has made significant progress by transforming multi-view image inputs into a unified BEV representation. Methods such as Lift-Splat-Shoot \cite{p8}, BEVFormer \cite{p9}, and PETR \cite{p10} leverage deep reasoning or attention interactions to map image features into BEV space. However, these methods generally assume that the images come from a pinhole camera and use linear projection, an assumption that does not hold for fisheye imaging systems. Furthermore, most existing methods ignore the unique geometric characteristics of surround-view fisheye systems, such as large overlap between adjacent viewpoints, significant radial distortion, and strong perspective differences, as shown in Fig. \ref{pin_fish}. Consequently, direct application of these methods to surround-view fisheye systems often results in performance degradation, spatial misalignment, and semantic inconsistency across views \cite{p5, p7, p11}.

\begin{figure}
    \centering
    \includegraphics[width=1\linewidth]{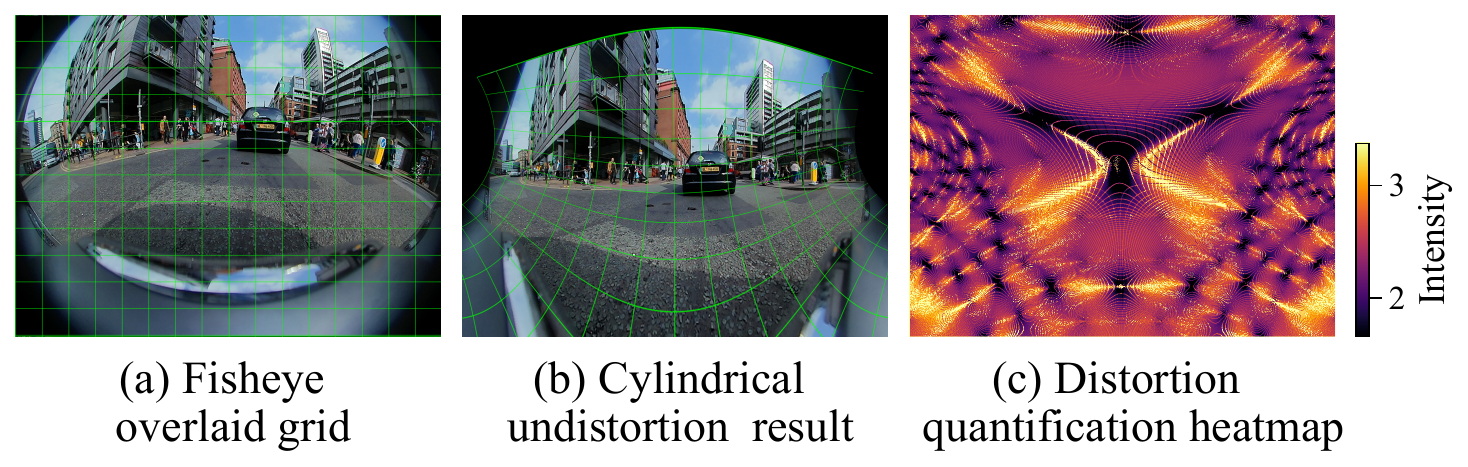}
    \caption{Distortion analysis of fisheye image. (a) Fisheye image with overlaid grid to intuitively show geometric distortion; (b) Cylindrical undistortion result to reveal limitations of traditional undistortion methods; (c) Anisotropy distortion quantification heatmap ($\log_{10}$ scale) for quantifying directional distortion distribution.}
    \label{fish_hot}
    \vspace{-0.8em}
\end{figure}

Despite the rapid development of BEV perception, research specifically targeting surround-view fisheye systems remains relatively scarce \cite{p12, p4}. As illustrated in Fig. \ref{fish_hot}, fisheye cameras suffer from severe radial distortions that deform grids near the periphery. Cylindrical projection can partially alleviate these effects, but noticeable inconsistencies remain \cite{p11}. A quantitative distortion heatmap further highlights non-uniform magnification and anisotropy, undermining the reliability of BEV features. These geometric and semantic challenges are further exacerbated by the ill-posed mapping from PV to BEV, making it difficult to develop, validate, and fairly benchmark specialized algorithms \cite{p15}. These challenges motivate our design of distortion- and uncertainty-aware BEV representations.

To address these challenges, a crucial step lies in learning robust fisheye representations that can generalize across diverse scenes, illumination conditions, and severe geometric distortions. Conventional backbones, such as ResNet \cite{p33} and VoVNet \cite{p34}, are typically trained on limited-scale datasets and often fail to capture distortion-resilient and semantically consistent features from fisheye images. In contrast, recent advances in large-scale self-supervised pretraining, particularly DINOv2 \cite{p35}, have demonstrated remarkable capability in extracting generic and transferable features across domains \cite{p36, p37, p38}. Motivated by this, we adopt DINOv2 as a multi-scale backbone to enhance fisheye feature extraction, providing a strong foundation for constructing accurate and robust BEV representations.

Although DINOv2 provides strong single-view features, the perspective distortion and overlapping fields of view in surround-view fisheye cameras introduce semantic ambiguity and feature uncertainty during multi-view fusion. Unlike pinhole images, fisheye projections are nonlinear and heavily distorted, leading to inconsistent or unreliable features when mapped to BEV space \cite{p4, p5}. Previous methods often fuse multi-view features by simple averaging or summation, ignoring their varying reliability. Such naive aggregation dilutes reliable signals and amplifies errors from ambiguous regions, resulting in degraded BEV quality. To tackle this problem, we introduce an Uncertainty-aware Spatial Cross-Attention module that estimates feature confidence and integrates information with reliability-guided weighting. This design enables more effective multi-view fusion, improving both spatial consistency and semantic robustness in the BEV representation.

Building on these BEV features, the next crucial step for BEV perception is to enable effective temporal dynamic modeling. Driving scenes exhibit strong near–far differences: nearby objects often undergo rapid motion and require fine-grained temporal modeling, whereas far field regions evolve more slowly but demand global stability to preserve context across frames. Conventional temporal aggregation schemes, however, typically treat all spatial locations and distances uniformly, which can lead to over-smoothing of near field dynamics or instability in far field context. To address this issue, we propose a Distance-aware Temporal Self-Attention mechanism that explicitly incorporates distance priors into temporal modeling, allowing the network to adaptively emphasize fine-grained motion in the near field while maintaining coherent global awareness in the far field.

In summary, our contributions are as follows:
\begin{itemize}
    \item We propose FishBEV, a surround-view fisheye BEV segmentation framework that integrates large-scale pretrained backbones with task-specific designs to better handle fisheye distortions and multi-view fusion.
    \item We introduce an Uncertainty-aware spatial cross-attention (U-SCA) module that estimates feature uncertainty and performs reliability-guided multi-view fusion, thereby improving spatial consistency and semantic robustness.
    \item We design a Distance-aware Temporal Self-Attention (D-TSA) mechanism that explicitly incorporates distance priors into temporal modeling, enabling adaptive handling of near field dynamics and far field context.
\end{itemize}

\section{RELATED WORK}

\subsection{BEV Perception from Multi-Camera Images}
Existing BEV perception methods can be roughly divided into four categories according to the PV to BEV feature transformation strategies. Early geometric-based methods used camera calibration and inverse perspective mapping to warp multi-view images into BEV space \cite{p16}, which is efficient but suffers from calibration sensitivity and the assumption of a flat ground \cite{p17, p18, p19}. Depth-based methods, such as Lift-Splat-Shoot \cite{p8}, estimate per-pixel depth distributions to lift features into 3D space before projection, providing greater geometric flexibility \cite{p20, p21, p22}. MLP-based approaches learn direct mappings from image to BEV features without explicit geometry, enabling end-to-end optimization but typically requiring large-scale data \cite{p23, p24, p25}. Recently, query-based methods have adopted a transformer architecture \cite{p26, p27}, where BEV queries focus on multi-view features, enabling richer spatial reasoning and temporal integration \cite{p9, p28, p10, p29}.

\begin{figure*}
    \centering
    \includegraphics[width=1\linewidth]{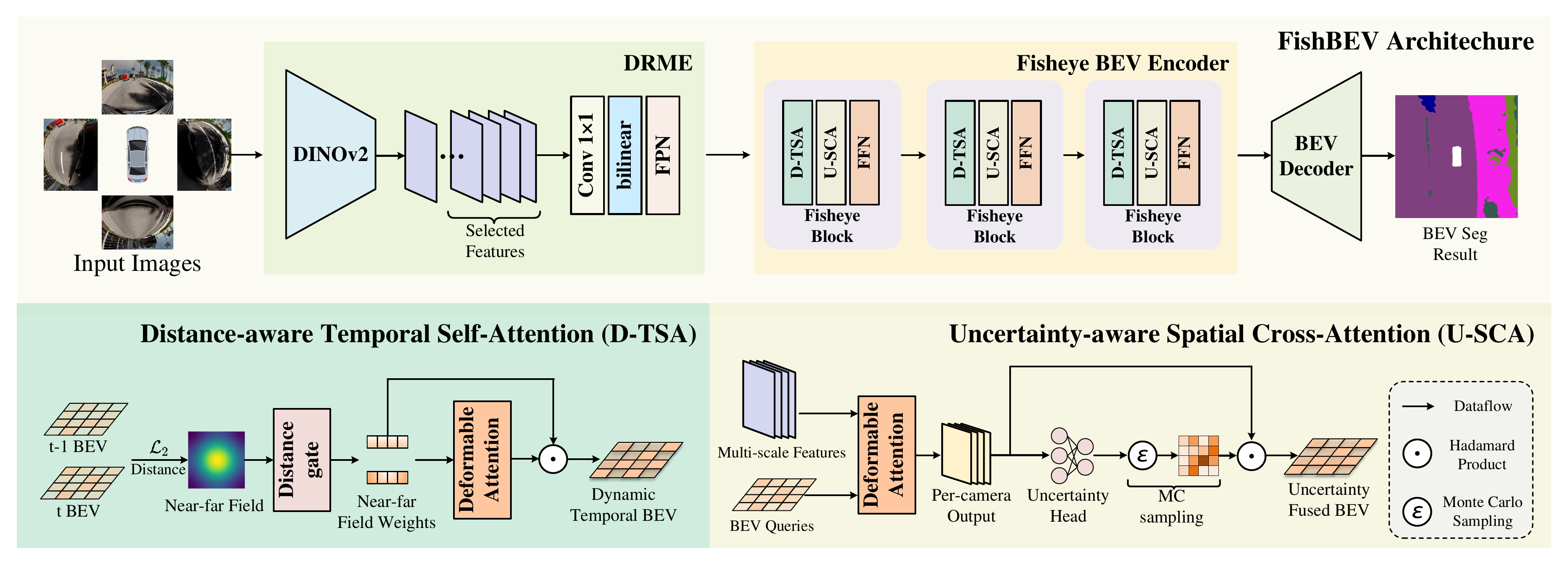}
    \caption{Overview of the proposed FishBEV framework. Surround-view fisheye images are encoded by the Distortion Resilient Multi-Scale Extraction (DRME) module, integrated with the Fisheye BEV Encoder consisting of Distance-Aware Temporal Self-Attention (D-TSA), Uncertainty-Aware Spatial Cross-Attention (U-SCA), and a Feedforward Network (FFN). Finally, the BEV decoder achieves qualified BEV segmentation results.}
    \label{FishBEV}
    \vspace{-0.8em}
\end{figure*}
\subsection{Surround-view Fisheye Camera Perception}
While BEV perception from multi-camera images has achieved remarkable progress, most existing approaches assume a pinhole projection model and thus are not directly applicable to surround-view fisheye systems with strong nonlinear distortions. To address this gap, F2BEV \cite{p13} represents the first attempt to directly construct BEV representations from surround-view fisheye images, mapping BEV queries to PV features via a fisheye transformer, but its application is limited to parking lots. Building on the Lift-Splat-Shoot \cite{p8} framework, FisheyeBEVSeg \cite{p14} introduces a distortion-aware learnable pooling strategy to better accommodate the nonlinear projection of fisheye images, but does not fully consider the distortion characteristics. ArticuBEVSeg proposes a flexible BEV segmentation framework tailored for articulated long combination vehicles, which integrates distorted fisheye images with time-varying extrinsics through implicit temporal alignment \cite{p30}. Other works explore geometric rectification or fisheye feature extraction to mitigate fisheye distortion, but research in this area is still limited, and challenges such as large field of view overlap, severe radial distortion, and balancing near-far field perception remain largely unaddressed \cite{p31, p32}.

\section{METHOD}

As illustrated in Fig. \ref{FishBEV}, the proposed FishBEV framework follows the standard image-to-BEV paradigm with a backbone, a BEV encoder, and a decoder. To handle fisheye-specific challenges, we redesign each stage with dedicated modules. At the backbone, the Distortion-Resilient Multi-scale Extraction (DRME) module captures robust multi-scale features under severe distortions. The BEV encoder incorporates Uncertainty-aware Spatial Cross-Attention (U-SCA) to refine multi-view alignment with uncertainty-guided fusion and Distance-aware Temporal Self-Attention (D-TSA) to balance near field detail and far field context for temporal coherence. Finally, a lightweight decoder produces dense BEV segmentation maps.

\subsection{Distortion-Resilient Multi-scale Extraction} 
The backbone of FishBEV is designed to capture robust multi-scale fisheye features while mitigating the negative impact of severe radial distortions. We build upon the strong semantic representation ability of DINOv2 \cite{p35}, and further enhance it with a multi-scale feature pyramid network (FPN) design \cite{p39}. Given a set of surround-view fisheye images, denoted as $X \in \mathbb{R}^{B\times C\times H\times W} $, where $B$ is the batch size, $C$ is the number of channels, and $H, W$ are image height and width respectively, we first partition the image into non-overlapping patches of size $P\times P$. These patches are linearly projected and fed into a pre-trained DINOv2, producing hidden representations $F=\left \{F_{i} \,|\,i \in [1,\cdots,L], F_{i}\in 
\mathbb{R}^{B\times (N+1) \times D} \right \}, $ at each transformer layer $l\in \left \{ 1,\dots ,L-1,L \right \}$:
\begin{equation}
    F = \mathrm{DINOv2}(\mathrm{PatchEmbed}(X)) ,
\end{equation}
where $N=\frac{H\cdot W }{P^{2}} $ is the number of patch tokens and D is the hidden dimension. We choose the output of the last four layers as the source of multi-scale features, i.e. $l\in \left \{ L-3,L-2,L-1,L \right \}$, to ensure global semantic consistency and preserve local details. We reshape the token sequence into a two-dimensional feature map $F^{\prime}_{l}\in \mathbb{R}^{B\times D\times H_{l} \times W_{l}}, H_{l}=H/P, W_{l}=W/P$:
\begin{equation}
    F^{\prime}_{l}=\mathrm{Reshape}(F_{l}[:,1:,:]),
\end{equation}
Subsequently, the features of each scale are channelized and mapped to a uniform number of channels through convolution, and bilinear upsampling or downsampling is performed according to the scale to obtain the multi-scale feature set $ F_{l}^{\prime \prime }=\left \{F^{\prime \prime}_{1},\cdots, F^{\prime \prime}_{L} \right \}$:
\begin{equation}
    F_{l}^{\prime \prime}=\mathrm{Interp}(\mathrm{Conv}(F^{\prime}_{l})), \quad F_{\mathrm{DRME}}=\mathrm{FPN}(\left \{ F_{l}^{\prime \prime} \right \} ).
\end{equation}

Finally, the distortion-resilient multi-scale feature representations $F_{\mathrm{DRME}}$ are obtained through FPN fusion. In summary, DRME effectively leverages the semantic richness of the final transformer layers while preserving spatial details through multi-scale fusion. This design not only enables robust representation of objects at varying scales but also alleviates the distortion bias inherent in fisheye cameras. As a result, DRME provides features that form a solid foundation for the subsequent BEV encoder.

\subsection{Fisheye BEV Encoder}

After extracting multi-scale fisheye features with DRME, we design a dedicated Fisheye BEV Encoder to further model temporal and spatial relationships, while explicitly accounting for geometric distortions and uncertainties inherent in fisheye imagery. The encoder is composed of multiple stacked Fisheye BEV Blocks, each consisting of three core modules: Uncertainty-aware Spatial Cross-Attention (U-SCA), Distance-aware Temporal Self-Attention (D-TSA) and a Feed-Forward Network (FFN). Specifically, U-SCA enables robust interaction between BEV queries and multi-view features via uncertainty modeling, D-TSA captures temporal dependencies across frames with distance-aware weighting and FFN enhances non-linear representation capacity. By stacking multiple blocks, the encoder progressively strengthens the spatial-temporal consistency and robustness of BEV representations, laying a solid foundation for downstream BEV segmentation task.

\subsubsection{Uncertainty-aware Spatial Cross-Attention}
Accurately lifting features from highly distorted fisheye images into a coherent BEV representation remains a central challenge in surround-view perception, due to the severe distortion and noise inherent in fisheye cameras \cite{p9}. Conventional spatial cross-attention (SCA), which projects BEV queries onto image features, is particularly sensitive to these effects. To address this, we propose an Uncertainty-aware Spatial Cross-Attention (U-SCA) module, which dynamically estimates the reliability of features from each camera view and fuses them using uncertainty-weighted aggregation, ensuring that the BEV representation emphasizes more trustworthy information.

In the U-SCA module, each BEV query $q_{i}\in \mathbb{R}^{C}$ at position $p_{i}=(x_{i}, y_{i})$ is projected onto all $N_{c}$ fisheye camera views using the fisheye projection model \cite{p11}. For a given camera $c$, this projection yields a 2D reference point $p_{i,c}$ on the feature map $V_{c}\in \mathbb{R}^{H\times W\times C}$. The corresponding deformable attention operation is formulated as \cite{p40}:
\begin{equation}
f_{i,c}=\sum_{k=1}^{K}A_{i,c,k}\cdot V_{c}(p_{i,c}+\bigtriangleup p_{i,c,k}),
\end{equation}
where $f_{i,c}\in \mathbb{R}^{C}$ denotes the camera-specific feature vector for query $q_{i}$, $A_{i,c,k}$ is the attention weight, and $\bigtriangleup p_{i,c,k}$ is the learnable sampling offset. Here, $i$ indexes the BEV query, $c$ indexes the camera view, and $k$ indexes the sampling point in deformable attention.

A key insight is that not all $f_{i,c}$ are equally reliable. Features from cameras whose projection points lie in heavily distorted regions should contribute less to the final output. To address this issue, we design a lightweight two-branch network consisting of a mean estimation head $\mathrm{Head}_{\mu}$ and a log-variance estimation head $\mathrm{Head}_{\sigma}$. Both heads are implemented as two-layer MLPs, and they jointly output the distribution parameters corresponding to each camera observation:
\begin{equation}
    \mu_{i,c} = \mathrm{Head}_{\mu}(f_{i,c}), \quad\log(\mathrm{var}_{i,c})=\mathrm{Head}_{\sigma}(f_{i,c}),
\end{equation}
where $\mu_{i,c},\,\mathrm{log}(\mathrm{var}_{i,c})$ denote the predicted mean and the logarithm of the variance, respectively. Using the reparameterization method \cite{p48}, we draw samples:
\begin{equation}
    z_{i,c}^{(s)}=\mu_{i,c}+\sigma_{i,c}\odot \epsilon^{(s)},
\end{equation}
\begin{equation}
    \sigma_{i,c}=\mathrm{exp}(\frac{1}{2}\log (\mathrm{var}_{i,c})), \ \epsilon^{(s)}\sim \mathcal{N} (0,1).
\end{equation}

To obtain an empirical measure of fusion uncertainty, we employ Monte Carlo (MC) sampling \cite{p42}, generating $K_{MC}$ samples per query. For each Monte Carlo iteration $s\in \left \{1,\cdots,K_{MC}\right \}$, per-camera samples are fused with precision-based weighting:
\begin{equation}
   \tilde{f}^{(s)}_{i}=\frac{\sum_{c}M_{i,c}w_{i,c}z^{(s)}_{i,c}}{\sum_{c}M_{i,c}w_{i,c}+\xi}, \quad w_{i,c}=\frac{1}{\mathrm{var}_{i,c}+\xi} ,
\end{equation}
with $M_{i,c}$ denoting the visibility mask and $\xi$ representing a small constant used in weighting to avoid division by zero. The final fused feature is obtained as the mean across Monte Carlo samples:
\begin{equation}
    \tilde{f}_{i}=\frac{1}{K_{MC}}\sum_{s=1}^{K_{MC}}\tilde{f}_{i}^{(s)},
\end{equation}
where $K_{MC}$ is the number of Monte Carlo sampling rounds. The sample variance of $\left \{ \tilde{f}_{i}^{(s)}\right \}_{s=1}^{K_{MC}}$ serves as a confidence metric for the fused feature, defined as:
\begin{equation}
\text{conf}_i = \frac{1}{K_{MC}-1} \sum_{s=1}^{K_{MC}} \left( \tilde{f}_i^{(s)} - \tilde{f}_i \right)^2,
\end{equation}
where a smaller $\mathrm{conf}_{i}$ indicates higher reliability of $\tilde{f}_{i}$. 

To normalize the uncertainty head and prevent variance collapse or divergence, we introduce a Kullback-Leibler (KL) divergence loss to regularize the distribution of prediction. This loss enforces the predicted uncertainty to align with a reasonable prior, ensuring stability during training and reliability of the uncertainty-aware fusion. The KL divergence loss is formulated as:
\begin{equation}
    \mathcal{L}_{KL}=\frac{1}{B\cdot N_{c}\cdot N_{q}}\sum_{b,i,c}\frac{1}{2}[\log \frac{\mathrm{var}_{prior}}{\mathrm{var}_{i,c}}+\frac{\mathrm{var}_{i,c}+\mu^{2}_{i,c}}{\mathrm{var}_{prior}}-1].
\end{equation}
Specifically, We define the prior distribution $\mathcal{N}\sim(0, \mathrm{var}_{prior})$  using $\log \mathrm{var}_{prior}=-4.0$. This prior balances small noise (avoiding feature distortion) and meaningful uncertainty (enabling reliability differentiation) for fisheye features. Thus, U-SCA enhances BEV feature construction by explicitly modeling per-camera uncertainty and fusing features based on their estimated reliability, leading to more robust and trustworthy representations in highly distorted fisheye scenarios.

\subsubsection{Distance-aware Temporal Self-Attention}

Conventional temporal self-attention treats all spatial positions equally when aggregating across frames \cite{p9, p13}. In fisheye imagery, however, near field features are typically more reliable than far field ones due to distortion and resolution loss. To this end, we propose D-TSA, which injects per-position distance information into the attention computation so that temporal fusion emphasizes near field contributions and suppresses noisy far field signals, improving temporal consistency and near field dynamic perception in BEV.

Given an input sequence of BEV queries $Q_{t} \in \mathbb{R}^{N_{q}\times C}$ at time step $t$, where $N_{q}$ represents the number of queries, and a short BEV queue $\Omega=\left\{ Q_{t-1}, Q_{t} \right\}$ containing historical and current BEV features, the D-TSA module computes attention using deformable sampling \cite{ p41} and distance-aware gating. Different from the deformable attention of global fixed temporal weight in BEVFormer, D-TSA injects spatial position information into temporal weight calculation through distance-aware gating, making the fusion strategy adaptive to the near field and far field characteristics of the fisheye camera. This allows flexible aggregation over both temporal and spatial dimensions while considering the relative importance of near and far field regions. For each query $q_{i}\in Q_{t}$, we sample $K$ spatial locations around reference points $p_{i}$ with learnable offsets $\bigtriangleup p_{i,k}^{\tau }$ for each BEV frame $\tau \in \left \{  t-1, t\right \}$:
\begin{equation}
    s_{i,k}^{\tau}=p_{i}+\bigtriangleup p_{i,k}^{\tau },
\end{equation}
where $i=1,\dots,N_{q}$ and $k=1,\dots,K$. Here, $s_{i,k}^{\tau}$ represents the sampling position, $p_{i}$ is calculated by combining the fisheye camera model with the camera's intrinsic and extrinsic matrix projections \cite{p11}. To modulate the contribution of current and historical BEV queries based on spatial location, for a query located at $p_{i}$, we compute its normalized Euclidean distance to the BEV center $O$:
\begin{equation}
    \bar{d}_{i}=\frac{\left \| p_{i}-O \right \|_{2} }{R} ,
\end{equation}
with $R=\frac{W}{2}$ representing the maximum BEV radius and smooth gating factor is then defined via a sigmoid function:
\begin{equation}
    \gamma_{i}=\sigma(\kappa(\delta-\bar{d}_{i})),
\end{equation}
where $\kappa$ controls the slope of the transition and $\delta\in [0,1]$ is the near-far threshold. Here, $\gamma_{i}\in [0, 1)$, the $\gamma_{i}$ of near field query approaches 1, and the $\gamma_{i}$ of far field query approaches 0, realizing dynamic weight distribution of near field focusing on current frame and far field focusing on historical frame. Based on the gating factor $\gamma_{i}$, we optimize the attention weight:
\begin{equation}
    \hat{w}_{i,k}^{(t-1)}=(1-\gamma_{i})w_{i,k}^{(t-1)}, \quad \hat{w}_{i,k}^{(t)}=\gamma_{i}w_{i,k}^{(t)},
\end{equation}
with $w_{i,k}^{(t-1)}$ and $w_{i,k}^{(t)}$ denoting the deformable attention weights for frames $t-1$ and $t$, respectively. Finally, the fused representation integrates both temporal features:
\begin{equation}
    \tilde{Q}_{i}  = \sum_{k=1}^{K}(\hat{w}_{i,k}^{(t-1)}Q_{t-1}(s_{i,k}^{t-1})+\hat{w}_{i,k}^{(t)}Q_{t}(s_{i,k}^{(t)})).
\end{equation}

This design allows the model to emphasize near field regions, which are more reliable due to higher resolution and lower distortion, while attenuating far field signals that are prone to noise. By fusing historical and current BEV features with spatially modulated attention weights, D-TSA improves temporal consistency, stabilizes dynamic object representation, and facilitates robust multi-frame aggregation in challenging fisheye scenarios.
% \begin{figure}
%     \centering
%     \includegraphics[width=1\linewidth]{Figure/TSA.pdf}
%     \caption{Enter Caption}
%     \label{fig:placeholder}
% \end{figure}

We retain the standard feed-forward network (FFN) module in each Fisheye BEV block, which is the same as BEVFormer  \cite{p9, p47}. It consists of a linear projection that expands the feature dimension from 256 to 512, followed by a ReLU activation and dropout regularization. A second linear layer projects the features back to 256 dimensions, with an additional dropout layer to mitigate overfitting. This module refines the intermediate BEV representations in a lightweight yet effective manner.

\subsection{BEV Decoder}

For the decoder we adopt the MaskHead~\cite{p43} to produce the final BEV segmentation masks. Concretely, each query is split into a content vector and a positional vector and fed into the MaskHead, which performs several decoding layers of query feature interaction. The decoder projects its outputs to low-resolution per-class BEV masks of shape $B\times C_{stuff}\times H_{BEV}\times W_{BEV}$. These masks are then upsampled by bilinear interpolation and optionally refined with small convolutional layers to match the target output resolution. Finally, the decoder produces high-resolution semantic masks after upsampling and refinement, which serve as the BEV segmentation output of FishBEV.

\subsection{Loss Function}
To address the class imbalance issue in BEV panoptic segmentation, we adopt a Focal Loss as the main task loss \cite{p44}. This loss down-weights easy examples and focuses on hard-to-classify pixels, which is critical for learning fine-grained details in fisheye BEV features. Let the model logits be $S\in \mathbb{R}^{B\times C_{stuff}\times H_{BEV}\times W_{BEV}}$ and the ground truth be $Y$. The task loss is denoted $\mathcal{L}_{focal}(\cdot)$. To regularize the uncertainty head of U-SCA and avoid variance collapse and divergence, we add the previously defined KL divergence term $\mathcal{L}_{KL}$ to the objective during training:
\begin{equation}
    \mathcal{L}_{total}=\mathcal{L}_{focal}(S,Y)+\lambda_{KL}\mathcal{L}_{KL}
\end{equation}
where $\lambda_{KL}$ controls the KL strength, we use $\lambda_{KL}=0.01$ by default.

\begin{figure*}
    \centering
    \includegraphics[width=0.8\linewidth]{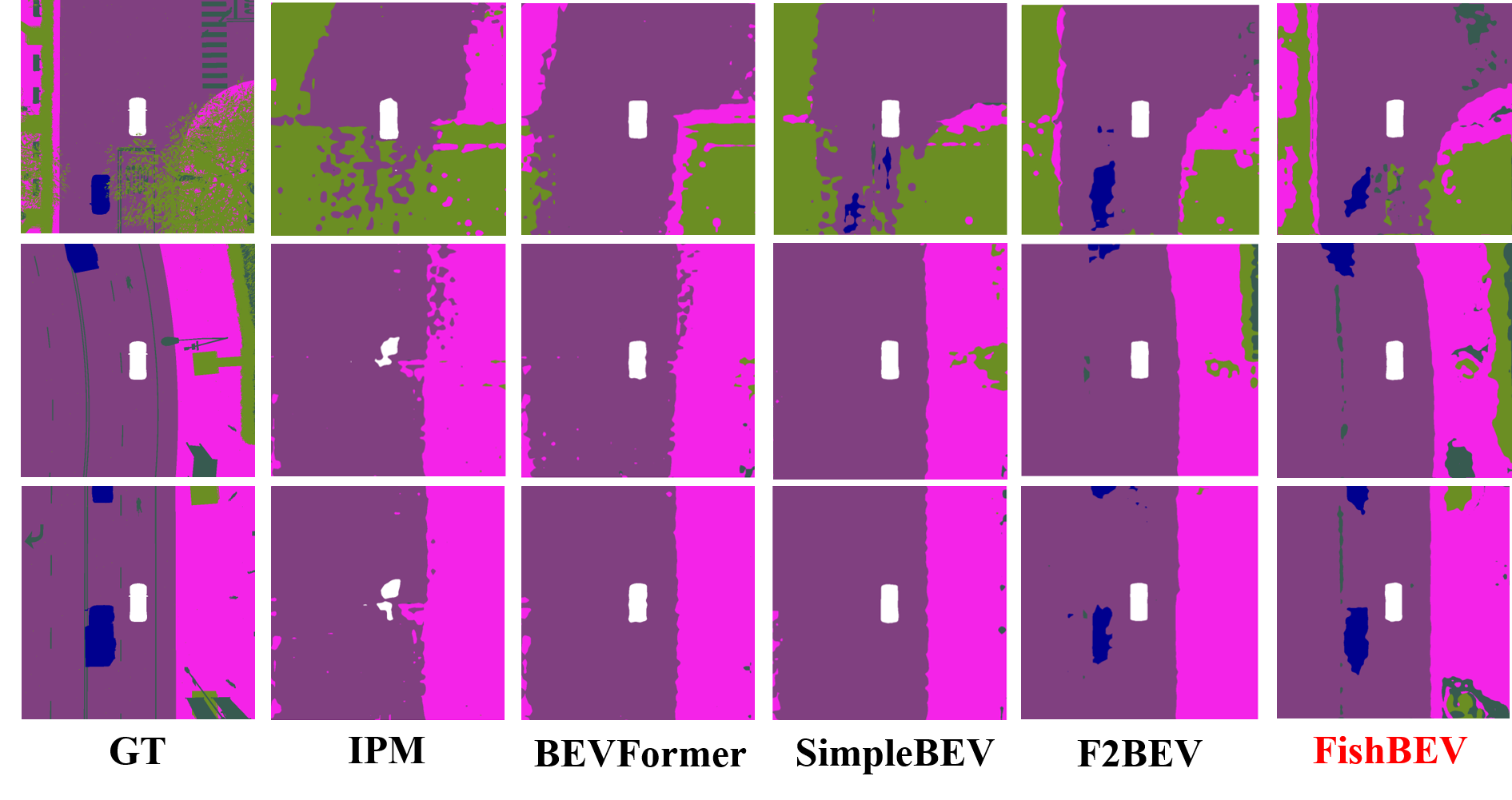}
    \caption{BEV segmentation results compared with baselines on the SynWoodscapes dataset.}
    \label{segresult}
    \vspace{-0.8em}
\end{figure*}
\section{Experiments}

\subsection{Experimental Settings}
\subsubsection{Dataset}We evaluate the proposed FishBEV model on the SynWoodscapes dataset \cite{p15}, a surround-view fisheye perception dataset built on the CARLA simulation platform \cite{p45}. The dataset contains surround-view fisheye images of resolution $1280\times966$, with corresponding BEV annotations of size $1024\times1024$. Each fisheye camera has a field of view of $190^{\circ}$, covering diverse road types and driving scenarios. SynWoodscapes defines 25 original semantic categories, including road structures, static objects, dynamic objects, and background classes. To incorporate the long-tail characteristic of SynWoodscapes and adapt the dataset to our BEV segmentation task, we remap the annotations into six core categories according to their proportions: void, road, sidewalk, vegetation, four-wheeler vehicle, and ego-vehicle. To ensure that the simulated fisheye images faithfully reproduce real-world distortions, SynWoodscapes employs a distortion polynomial model:
\begin{equation}
    r(\theta)=a_1\theta+a_2\theta^2+a_3\theta^3+a_4\theta^4,
\end{equation}
which is combined with cubemap projection and ray-tracing techniques. The coefficients ${a_i}$ control the strength of different orders of distortion, while $\theta$ denotes the incident angle between the incoming ray and the optical axis. Moreover, it incorporates various environmental perturbations such as fog, rain, and illumination changes, enabling robust evaluation of model generalization under diverse conditions.
\begin{table}[t]
    \centering
    \caption{Trainable parameter counts and mIoU for different DRME backbones. 
    ``Params.'' denotes trainable parameters in millions (MB).}
    \label{params}
    \begin{tabular}{lcc}
        \toprule
        \textbf{Backbone (FishBEV)} & \textbf{Trainable Params. (MB)}\\
        \midrule
        ResNet34-FPN (baseline)     & 23.89 \\
        DINOv2-Small-FPN (D-S)      & \textbf{12.60} \\
        DINOv2-Base-FPN (D-B)       & 33.21 \\
        DINOv2-Large-FPN (D-L)      & 68.59 \\
        \bottomrule
    \end{tabular}
    \vspace{-0.8em}
\end{table}
\subsubsection{Evaluation Metrics}
We evaluate model performance using the mean Intersection-over-Union (mIoU), a standard metric for semantic segmentation tasks. For each class $c$, the Intersection-over-Union (IoU) is defined as:
\begin{equation}
\mathrm{IoU}_c = \frac{TP_c}{TP_c + FP_c + FN_c},
\end{equation}
where $TP_c$, $FP_c$, and $FN_c$ denote the number of true positive, false positive, and false negative pixels for class $c$, respectively. The final mIoU is obtained by averaging across all $C$ valid semantic classes:
\begin{equation}
\mathrm{mIoU} = \frac{1}{C} \sum_{c=1}^{C} \mathrm{IoU}_c.
\end{equation}

In our experiments, $C=6$ after remapping the original 25 categories into six core classes. The void class is excluded from evaluation. mIoU effectively measures both the accuracy of pixel-wise classification and the balance across categories, making it particularly suitable for assessing BEV semantic segmentation under long-tailed distributions.

\begin{table}[t]
    \centering
    \caption{Freezing strategy for DINOv2 in the DRME backbone. D-S, D-B, and D-L represent the small, base, and large versions of the DINOv2 model, respectively.}
    \begin{tabular}{lcc}
        \toprule
        \textbf{Model Type} & \textbf{Freeze Ratio} & \textbf{Selected Layers} \\
        \midrule
        D-S  &  60\% & [8, 9, 10, 11]  \\
        D-B   &  70\% & [8, 9, 10, 11]  \\
        D-L  &  80\% & [20, 21, 22, 23] \\
        \bottomrule
    \end{tabular}
    \label{freeze}
    \vspace{-0.8em}
\end{table}

\subsubsection{Implementation Details}
We implemented FishBEV using PyTorch with distributed data-parallel training on two NVIDIA A6000 GPUs. The BEV queries size is setting to $50\times50$. The input surround-view fisheye images were resized to a resolution of $640\times540$, along with the can bus status data of vehicle. Data augmentation included random color dithering and gamma correction. Each training sample consisted of a sequence of three consecutive frames. The freezing strategy of backbone is shown in Table \ref{freeze}. Feature extraction was then performed using the proposed fisheye BEV encoder. The near-far field threshold coefficient $\delta$ is set to 0.8, and the slope $\kappa$ is set to 10. We used AdamW as the optimizer with an initial learning rate of $3\times10^{-5}$ and a decay of 0.99 per epoch. The model was trained for 50 epochs with a batch size of 2 per GPU.

\begin{table*}[t]
    \centering
    \caption{Comparison with baselines on the SynWoodscapes dataset (IoU in \%). The best results are highlighted in bold.}
    \begin{tabular}{lcccccc}
        \toprule
        \textbf{Method} & \textbf{Road} & \textbf{Sidewalk} & \textbf{Vegetation} & \textbf{Four-wheeler-vehicle} & \textbf{Ego-vehicle} & \textbf{mIoU} \\
        \midrule
        IPM             & 69.78 & 44.31 & 1.09 & 1.55 & 87.74 & 40.89 \\
        BEVFormer       & 82.64 & 59.69 & 8.83 & 4.46 & 91.27 & 49.37 \\
        SimpleBEV       & 81.34 & 59.27 & 6.62 & 8.96 & 91.29 & 49.49 \\
        F2BEV           & 85.68 & 62.15 & 15.59 & 8.13 & 95.42 & 53.39 \\
        \midrule
        FishBEV (D-S)   & 89.81 & 72.61 & 4.52 & 15.43 & 94.67 & 55.41 \\
        FishBEV (D-B)   & 91.20 & 75.92 & 21.85 & 27.58 & 93.75 & 62.06 \\
        FishBEV (D-L)   & \textbf{91.90} & \textbf{81.73} & \textbf{22.32} & \textbf{28.67} & \textbf{96.49} & \textbf{64.22} \\
        \bottomrule
    \end{tabular}
    \label{baseline_results}
\end{table*}

\begin{table*}[t]
\centering
\caption{Ablation study of FishBEV on the SynWoodscapes dataset (IoU in \%, best results in bold).}
\label{ablation}
\begin{tabular}{lcccccc}
\toprule
\textbf{Method} & \textbf{Road} & \textbf{Sidewalk} & \textbf{Vegetation} & \textbf{Four-wheeler-vehicle} & \textbf{Ego-vehicle} & \textbf{mIoU} \\
\midrule
Baseline & 90.17 & 69.34 & 8.05 & 12.78 & 91.60 & 54.39 \\
+ DRME & 91.01 & 74.92 & 17.24 & 18.71 & 94.37 & 59.65 \\
+ DRME + U-SCA & 91.11 & 79.12 & 19.97 & 24.70 & 95.43 & 62.07 \\
+ DRME + U-SCA + D-TSA & \textbf{91.90} & \textbf{81.73} & \textbf{22.32} & \textbf{28.67} & \textbf{96.49} & \textbf{64.22} \\
\bottomrule
\end{tabular}
\vspace{-0.8em}
\end{table*}

\subsection{Comparison with Baselines}
We compare our FishBEV model with several representative BEV perception methods, including Inverse Perspective Mapping (IPM) \cite{p16}, BEVFormer \cite{p9}, SimpleBEV \cite{p46}, and F2BEV \cite{p13}. The experimental results are summarized in Table~\ref{baseline_results}, and the BEV segmentation visualization results are shown in Fig. \ref{segresult}. Compared with existing baselines, FishBEV achieves consistent improvements across all categories. Traditional IPM struggles due to its purely geometric projection, leading to extremely poor segmentation of vegetation (1.09) and four-wheeler vehicles (1.55), which highlights the limitations of non-learning-based methods. BEVFormer and SimpleBEV provide stronger baselines, achieving around 49.3–49.5 mIoU, yet their performance remains constrained in fisheye settings, particularly for small or dynamic objects such as vehicles. F2BEV improves upon them with a fisheye-specific design, reaching 53.4 mIoU, but its encoder is still inherited from pinhole-camera assumptions, which makes it difficult to fully exploit fisheye-specific cues.

In contrast, it can be seen from Tables \ref{params} and \ref{baseline_results} that FishBEV shows obvious advantages in both accuracy and parameter efficiency. With DINOv2-Small (D-S), our model already surpasses F2BEV at 55.39 mIoU using only 12.6M parameters—nearly half the size of ResNet34. Scaling further boosts performance: FishBEV (D-B) reaches 62.06 mIoU with 33.2M parameters, showing strong gains on vegetation (+17.3) and vehicles (+12.2). Finally, FishBEV(D-L) achieves the best overall performance with \textbf{64.22} mIoU at 68.6M parameters. Notably, the model achieves substantial gains on sidewalk (+19.6 compared with F2BEV) and four-wheeler vehicles (+20.5), demonstrating its ability to capture both structural and instance-level semantics under severe fisheye distortions. These results highlight that FishBEV’s improvements are largely attributed to its architectural designs rather than merely scaling trainable parameters.

\subsection{Ablation Study}
To validate the effectiveness of each proposed component, we conduct ablation studies, as shown in Table~\ref{ablation}. Starting from the baseline model, which employs a plain transformer encoder without fisheye-specific designs, we progressively add the proposed modules. Incorporating the DRME improves mIoU from 54.39 to 59.65, with clear gains on sidewalk and vegetation, demonstrating the importance of distortion-resilient multi-scale extraction. Building upon this, the introduction of U-SCA further boosts mIoU to 62.07, mainly enhancing vegetation and four-wheeler vehicle categories, indicating that uncertainty-aware cross-attention effectively alleviates fisheye distortions and projection noise. Finally, adding the D-TSA yields the best overall performance of 64.22 mIoU. D-TSA brings consistent improvements across all categories, especially on sidewalk and vehicle classes, showing the benefits of distortion-invariant multi-scale enhancement. Overall, each module contributes positively, and their combination leads to a substantial performance gain of \textbf{9.83} mIoU over the baseline.

\section{CONCLUSIONS}
In this work, we introduced FishBEV, a novel framework for fisheye-to-BEV semantic segmentation that integrates Distortion-Resilient Multi-scale Extraction, Uncertainty-aware Spatial Cross-Attention and Distance-aware Temporal Self-Attention. These components address the challenges of severe fisheye distortion, near-far field perception discrepancies and temporal inconsistencies, enabling FishBEV to achieve significant improvements over existing baselines. Ablation studies further verify that each module contributes complementary improvements, confirming the effectiveness of our fisheye-aware design. In the future, FishBEV can be extended to more diverse autonomous driving scenarios, and its general design also holds promise for broader perception tasks such as 3D object detection and occupancy prediction in surround-view fisheye systems.

%\addtolength{\textheight}{-12cm}   % This command serves to balance the column lengths
                                  % on the last page of the document manually. It shortens
                                  % the textheight of the last page by a suitable amount.
                                  % This command does not take effect until the next page
                                  % so it should come on the page before the last. Make
                                  % sure that you do not shorten the textheight too much.

%%%%%%%%%%%%%%%%%%%%%%%%%%%%%%%%%%%%%%%%%%%%%%%%%%%%%%%%%%%%%%%%%%%%%%%%%%%%%%%%

%%%%%%%%%%%%%%%%%%%%%%%%%%%%%%%%%%%%%%%%%%%%%%%%%%%%%%%%%%%%%%%%%%%%%%%%%%%%%%%%

%%%%%%%%%%%%%%%%%%%%%%%%%%%%%%%%%%%%%%%%%%%%%%%%%%%%%%%%%%%%%%%%%%%%%%%%%%%%%%%%

%%%%%%%%%%%%%%%%%%%%%%%%%%%%%%%%%%%%%%%%%%%%%%%%%%%%%%%%%%%%%%%%%%%%%%%%%%%%%%%%

\bibliographystyle{IEEEtran}
\bibliography{main}

\end{document}